\documentclass[conference]{IEEEtran}
\IEEEoverridecommandlockouts
\usepackage{cite}
\usepackage{amsmath,amssymb,amsfonts}
\usepackage{graphicx}
\usepackage{amsmath}
\usepackage{algorithm}
\usepackage{algpseudocode}
\usepackage{textcomp}
\usepackage{tabularx}
\usepackage{booktabs}  
\usepackage{enumitem} 
\usepackage{xcolor}
\usepackage{caption}
\usepackage{geometry}
\usepackage{tikz}
\usepackage{subcaption}
\usetikzlibrary{shapes, arrows.meta,shapes.geometric, arrows, positioning, calc}
\usepackage{pgfplots}
\pgfplotsset{compat=1.18}
\usepackage{booktabs}
\usepackage{multirow}
\usepackage{lmodern}
\pgfplotsset{compat=newest}

\algnewcommand{\LineComment}[1]{\State \(\triangleright\) #1}
\def\BibTeX{{\rm B\kern-.05em{\sc i\kern-.025em b}\kern-.08em
    T\kern-.1667em\lower.7ex\hbox{E}\kern-.125emX}}
\begin{document}

\title{Agentic Trust Coordination for Federated Learning through Adaptive Thresholding and Autonomous Decision Making in Sustainable and Resilient Industrial Networks\\
}
\author{
\IEEEauthorblockN{\textbf{Paul Shepherd\IEEEauthorrefmark{1,2}, Tasos Dagiuklas\IEEEauthorrefmark{1}, Bugra Alkan\IEEEauthorrefmark{1}, Jonathan Rodriguez\IEEEauthorrefmark{2}}}
\IEEEauthorblockA{
\IEEEauthorrefmark{1} London South Bank University, London, UK. \\
\IEEEauthorrefmark{2} Instituto de Telecomunicações, Aveiro, Portugal. \\
paul@av.it.pt, tdagiuklas@lsbu.ac.uk, alkanb@lsbu.ac.uk, jonathan@av.it.pt}
}
\maketitle
\begin{abstract}
Distributed intelligence in industrial networks increasingly integrates sensing, communication, and computation across heterogeneous and resource constrained devices. Federated learning (FL) enables collaborative model training in such environments, but its reliability is affected by inconsistent client behaviour, noisy sensing conditions, and the presence of faulty or adversarial updates. Trust based mechanisms are commonly used to mitigate these effects, yet most remain statistical and heuristic, relying on fixed parameters or simple adaptive rules that struggle to accommodate changing operating conditions.

This paper presents a lightweight agentic trust coordination approach for FL in sustainable and resilient industrial networks. The proposed Agentic Trust Control Layer operates as a server side control loop that observes trust related and system level signals, interprets their evolution over time, and applies targeted trust adjustments when instability is detected. The approach extends prior adaptive trust mechanisms by enabling context aware intervention decisions, rather than relying on fixed or purely reactive parameter updates. By explicitly separating observation, reasoning, and action, the proposed framework supports stable FL operation without modifying client side training or increasing communication overhead.
\end{abstract}
\begin{IEEEkeywords}
Federated learning; Trust Management; Agentic AI; Adaptive control; ATSSSF; Agent Driven Coordination.
\end{IEEEkeywords}
\section{Introduction}
Industrial networks increasingly incorporate distributed intelligence to support monitoring, control, and optimisation across geographically and operationally dispersed assets. Learning based methods are commonly deployed at the network edge, where sensing quality, computational capability, and communication reliability vary across devices. In such environments, maintaining stable and dependable learning behaviour over time is a key requirement for both operational continuity and long term system sustainability.

Federated learning has emerged as a practical approach for training shared models in industrial settings where data centralisation is infeasible or undesirable due to privacy, regulatory, or operational constraints. By keeping training data local and aggregating model updates at a coordinating server, FL reduces data exposure and communication overhead. However, real deployments remain sensitive to inconsistent client behaviour, intermittent participation, non identical data distributions, and variations in data quality. These factors affect both global model accuracy and the stability of the training process.

Trust based mechanisms have been introduced to address these challenges by evaluating the reliability of client updates and regulating their influence on the global model, complementing aggregation rules and other robustness techniques. Existing approaches typically compute trust scores from statistical properties such as update similarity, participation consistency, or temporal behaviour. While such mechanisms can improve robustness, they often depend on fixed thresholds or manually defined adaptation rules. As operating conditions evolve, these static or weakly adaptive designs can lead to delayed responses, excessive exclusion, or oscillatory trust behaviour, which is undesirable in long lived industrial systems.

Recent implementations of our earlier work [1,2], the Adaptive Trust Score Scaling System Filtering (ATSSSF) which combined the TOPSIS [1,3] ranking method with EMA smoothing [1] to identify and limit unreliable contributions (weights), has shown that fixed thresholds and constant smoothing parameters restrict ATSSSF’s flexibility under dynamically changing client reliability conditions, and that dynamically adjusting trust parameters can improve stability under changing conditions. Nevertheless, these methods still treat adaptation as a parameter tuning problem rather than a control problem. Decisions are commonly triggered by local metric deviations without an explicit interpretation of overall training state. As a result, adaptation remains largely reactive and may fail to distinguish transient disturbances from sustained instability.

This paper addresses this limitation by introducing an agentic trust control perspective for FL [4]. Instead of embedding adaptation logic directly into trust update rules, the proposed approach introduces a lightweight server side control loop that observes trust related and system level signals, interprets their evolution over time, and selects appropriate trust control actions. By explicitly separating observation, reasoning, and action, the approach enables context aware intervention while preserving simplicity, interpretability, and deployability.

The remainder of this paper is organised as follows. Section II discusses background concepts and limitations of existing trust based FL approaches. Section III introduces the system model and design considerations. Section IV presents the proposed agentic trust control methodology. Section V discusses expected behaviour and implications for sustainability and resilience in industrial networks. Section VI concludes the paper and outlines directions for future work.
\section{Background and Related Work}
Federated learning (FL) enables collaborative model training across distributed clients while preserving data locality, making it attractive for industrial and edge environments. However, heterogeneity in data distributions, device capabilities, communication reliability, and participation patterns introduces instability that degrades convergence and model reliability [5,6]. Classical aggregation schemes such as Federated Averaging (FedAvg) and its extensions (e.g., FedProx) [7,8] address system heterogeneity but remain trust agnostic, implicitly assuming benign and reliable participants. In realistic industrial deployments, this assumption is increasingly invalid.

To address robustness limitations, trust aware FL frameworks associate each client with a trust score that evolves across training rounds. Trust scores are typically derived from structured behavioural signals, including update similarity, training stability, participation consistency, and performance indicators. By weighting or omitting client contributions based on trust, these approaches introduce temporal memory into the aggregation process and enable longer term regulation of unreliable or adversarial behaviour.

The ATSSSF framework [1,2] represents a lightweight and interpretable instantiation of this paradigm. ATSSSF employs multi-criteria decision making using TOPSIS with entropy based weighting to rank clients, while Exponential Moving Average (EMA) smoothing introduces temporal awareness. This design enables efficient mitigation of data poisoning and label flipping attacks without increasing communication overhead or modifying client-side training [1,2]. Its serverside operation and compatibility with frameworks such as Flower [9] make it suitable for MEC and industrial IoT deployments.

Despite these advantages, ATSSSF encodes adaptation through predefined operational rules. Omission thresholds and EMA parameters are manually specified and adjusted reactively in response to local metric deviations. In highly dynamic environments, such rule-based adaptation can lead to delayed responses, oscillatory client inclusion, or insufficient discrimination between transient noise and sustained behavioural drift. Similar limitations are observed in other adaptive trust and zero-trust aggregation schemes, which either rely on static control logic or incur prohibitive computational overhead [11].

Agentic and autonomic computing frameworks offer an alternative perspective by framing adaptation as a control problem rather than a parameter tuning task [10]. In these frameworks, system level signals are jointly observed, interpreted to infer an operating state, and used to select context dependent control actions through an explicit perception reasoning action loop. Lightweight agentic designs emphasise determinism, interpretability, and bounded decision logic, aligning well with the requirements of industrial and safety critical systems.

This work builds directly on ATSSSF by introducing an Agentic Trust Control Loop (ATCL) that supervises trust adaptation rather than redefining trust metrics or aggregation rules [11,12]. ATCL retains the structured trust signals and efficiency of ATSSSF while replacing predefined adaptation rules with an explicit observation–analysis decision action cycle. By inferring system states from joint trust and performance signals, ATCL enables context aware trust regulation that distinguishes between transient disturbances and sustained instability, representing a natural evolution from adaptive trust scaling to agent supervised trust control in FL.
\section{System Model and Design Goals}
We consider a synchronous FL system in which a central coordination server manages collaborative model training across a set of distributed industrial clients. Each client performs local optimisation on private data and periodically transmits model updates to the server, which aggregates the received updates to produce a global model that is redistributed in subsequent rounds. The system operates under partial and time varying participation, reflecting realistic industrial and MEC deployments characterised by heterogeneous computational resources, non-identically distributed data, variable sensing quality, and intermittent connectivity. The server is assumed to be trusted, while no assumptions are made regarding the reliability of individual clients.

In this setting, client behaviour may deviate from expected norms due to benign factors such as sensor noise, data drift, resource constraints, and communication disruptions, as well as adversarial actions including data poisoning or model manipulation. These effects are observable at the server through structured signals, including abnormal gradient patterns, elevated update volatility, inconsistent participation across rounds, and degradation in global loss trends. Existing State-Of-The-Art (SOTA) trust aware FL mechanisms utilise subsets of these signals to regulate aggregation; however, they typically rely on reactive, metric-specific rules and do not explicitly interpret system level behaviour across time.

The proposed ATCL is introduced as a lightweight, server-side supervisory mechanism that regulates trust adaptation prior to aggregation. ATCL does not modify client side training procedures, local optimisation objectives, or communication protocols. Instead, it operates on behavioural and performance signals already available at the server, enabling straightforward integration with existing FL frameworks. This positioning allows ATCL to complement, rather than replace, established trust aware aggregation mechanisms.

The design of ATCL is motivated by limitations in current adaptive trust approaches, which encode adaptation through predefined update rules and static control logic. While such methods improve robustness compared to fixed threshold baselines, they remain reactive and may respond similarly to transient noise and sustained behavioural degradation. ATCL moves Beyond this State-Of-The-Art (BSOTA) by framing trust adaptation as a control problem, in which multi-dimensional signals are jointly interpreted to infer operating conditions and select context-appropriate control actions.

Accordingly, ATCL is designed to enhance resilience by mitigating unreliable or adversarial client influence without destabilising training, to support sustainability by limiting unnecessary computation, communication overhead, and oscillatory client exclusion, and to remain deployable through lightweight, interpretable, and deterministic control logic. These design goals align ATCL with the practical requirements of sustainable and resilient industrial FL systems targeted by this workshop.
\section{methodology}
The Agentic Trust Control Loop (ATCL) extends adaptive ATSSSF mechanisms [1] by replacing predefined and locally reactive adaptation rules with a continuous observation analysis decision–action control cycle executed at the FL server. While ATSSSF introduced adaptive thresholding and temporal trust evolution, its adaptation logic remains rule driven and parameter centric. ATCL generalises this approach by introducing an explicit supervisory layer that coordinates trust adaptation based on inferred system states rather than direct metric responses.

The control loop begins by modelling a set of trust relevant signals that characterise client behaviour across federated rounds. These signals include statistical trust scores derived from gradient similarity patterns, volatility indicators capturing instability in local updates, similarity metrics comparing individual client behaviour to the federation’s normative profile, participation consistency measures reflecting long term reliability, and model level indicators such as global loss trends or anomalous convergence behaviour. Rather than treating these indicators as independent heuristics, ATCL integrates them into a structured and interpretable agent state representing the current health of the FL process.

Based on the inferred agent state, a lightweight reasoning layer executes the analysis phase of the control loop. For the moment, this reasoning component employs rule based or minimally agentic decision logic and is deliberately non–LLM-based, operating over a low dimensional, structured state representation derived from aggregated trust and system signals. By jointly evaluating multiple trust and system indicators over time, the agent detects emerging behavioural patterns, sustained deviations from nominal operation, and recovery trends. This enables ATCL to distinguish transient noise from persistent instability or adversarial behaviour, addressing a key limitation of threshold driven trust adaptation.

When the reasoning phase determines that intervention is required, ATCL issues targeted and state conditioned trust control actions. These actions include dynamically steering omission thresholds, adapting exponential moving average (EMA) smoothing factors governing trust score evolution, temporarily excluding unstable or low trust clients, and progressively reinstating previously excluded participants once stability is restored. In this manner, ATCL preserves the computational simplicity and server side deployability of ATSSSF while enabling more context aware, real time, and fine grained trust coordination.
\begin{table*}[t]
\caption{Comparison of Adaptive ATSSSF and Agentic Trust Control Loop (ATCL)}
\label{tab:atsssf_atcl}
\centering
\renewcommand{\arraystretch}{1.25}
\setlength{\tabcolsep}{8pt}
\begin{tabular}{p{3.6cm} p{5.6cm} p{5.6cm}}
\toprule
\textbf{Dimension} &
\textbf{Adaptive ATSSSF} &
\textbf{Agentic Trust Control Loop (ATCL)} \\
\midrule
Trust management paradigm &
Adaptive statistical trust filtering with parameter tuning &
Agent-supervised closed loop trust coordination \\

Adaptation logic &
Predefined rules governing threshold and EMA updates &
State conditioned decision making based on inferred system behaviour \\

Control structure &
Direct metric to parameter reaction &
Explicit observation--analysis--decision--action cycle \\

Use of temporal context &
Implicit via EMA smoothing &
Explicit multi-round trend interpretation and state inference \\

Signal integration &
Primarily independent trust indicators &
Joint reasoning over multiple trust and system-level signals \\

Decision trigger &
Local threshold violations or statistical deviation &
Inferred system state (normal, degraded, stabilising) \\

Handling of transient noise &
Reactive; may induce oscillatory trust updates &
Selective intervention distinguishing noise from sustained instability \\

Client exclusion strategy &
Threshold-driven omission &
Temporary, state conditioned exclusion and controlled reinstatement \\

Trust score evolution &
Parameter regulated smoothing &
Agent coordinated, context aware regulation \\

Interpretability &
Moderate (rule transparency) &
High (explicit agent state and control logic) \\

Computational overhead &
Low (server-side statistical processing) &
Low (lightweight reasoning over aggregated signals) \\

Communication overhead &
None beyond baseline FL &
None beyond baseline FL \\

Suitability for industrial FL &
Robust under moderate dynamics &
Designed for highly dynamic, heterogeneous industrial environments \\
\bottomrule
\end{tabular}
\end{table*}
\begin{algorithm}[t]
\caption{Agentic Trust Control Loop (ATCL)}
\label{alg:atcl}
\begin{algorithmic}[1]

\State Initialise trust scores $T_i$ for all clients $i$
\State Initialise omission threshold $\theta$
\State Initialise EMA smoothing factor $\alpha$
\State Initialise agent state $S \leftarrow \textsc{Normal}$

\For{each federated round $r$}

    \State Receive local model updates $\Delta w_i^r$ from participating clients

    \For{each client $i$}
        \State Compute update similarity $s_i^r$
        \State Compute volatility indicator $v_i^r$
        \State Update participation consistency $p_i^r$
        \State Update trust score $T_i^r$ using $\alpha$
    \EndFor

    \State Observe system level indicators:
    \State \quad global loss trend $L^r$
    \State \quad trust dispersion $\sigma_T^r$

    \Comment{Agentic analysis phase}
    \If{instability detected $(T_i^r, v_i^r, L^r)$}
        \State $S \leftarrow \textsc{Degraded}$
    \ElsIf{recovery detected $(T_i^r, L^r)$}
        \State $S \leftarrow \textsc{Stabilising}$
    \Else
        \State $S \leftarrow \textsc{Normal}$
    \EndIf

    \Comment{Decision and action phase}
    \If{$S = \textsc{Degraded}$}
        \State Adapt omission threshold $\theta$
        \State Adapt EMA factor $\alpha$
        \State Temporarily exclude low-trust clients
    \ElsIf{$S = \textsc{Stabilising}$}
        \State Gradually relax $\theta$ and $\alpha$
        \State Reinstate previously excluded clients
    \EndIf

    \State Aggregate updates using trust aware aggregation
    \State Broadcast updated global model
\EndFor
\end{algorithmic}
\end{algorithm}
\begin{figure}
    \centering
    \includegraphics[width=1.0\linewidth]{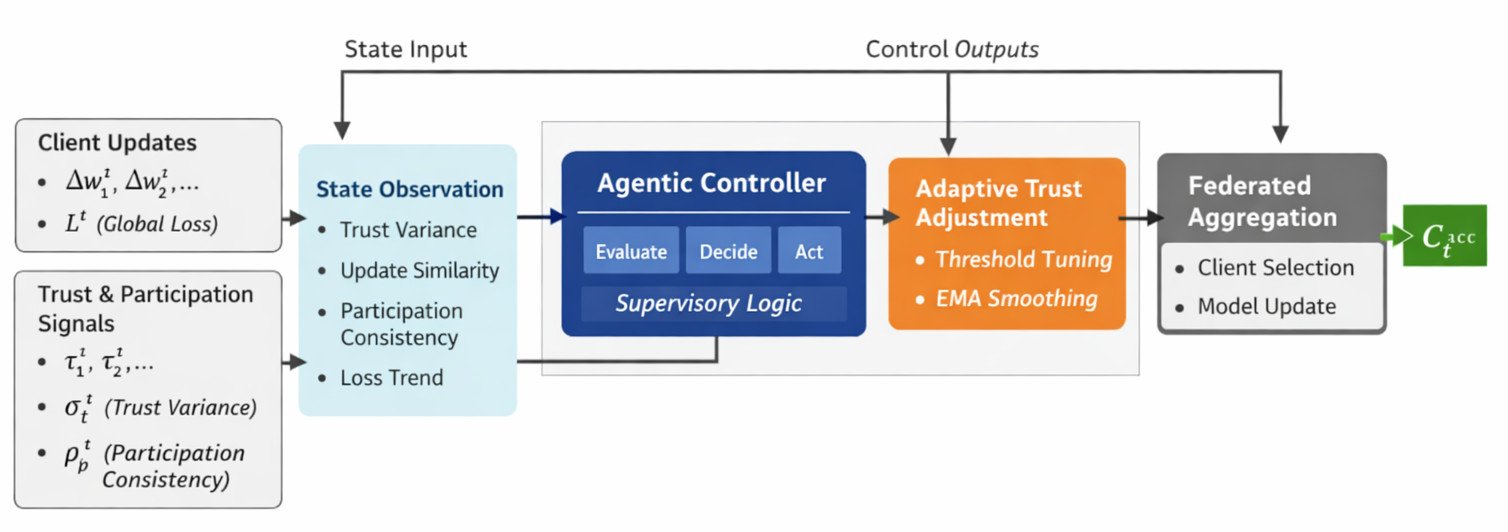}
    \caption{The Agentic Trust Control Loop (ATCL) integrated into the FL process.}
    \label{fig:placeholder}
\end{figure}
\section{Expected Results}
Although full experimentation is planned for later stages, the expected behaviour of the proposed framework is informed by earlier adaptive FL results obtained using ATSSSF [1,2]. In those studies, adaptive thresholding and adaptive EMA smoothing improved omission efficiency, reduced instability during adversarial periods, and produced more stable global model performance than fixed parameter baselines. Building on these findings, the ATCL is expected to provide additional improvements by introducing agentic behaviour [3,4] into the trust pipeline. The agent’s ability to observe evolving system signals, analyse behavioural trends, and autonomously select interventions allows trust regulation to shift from reactive adjustment to proactive pattern recognition.
\begin{figure}
    \centering
    \includegraphics[width=1\linewidth]{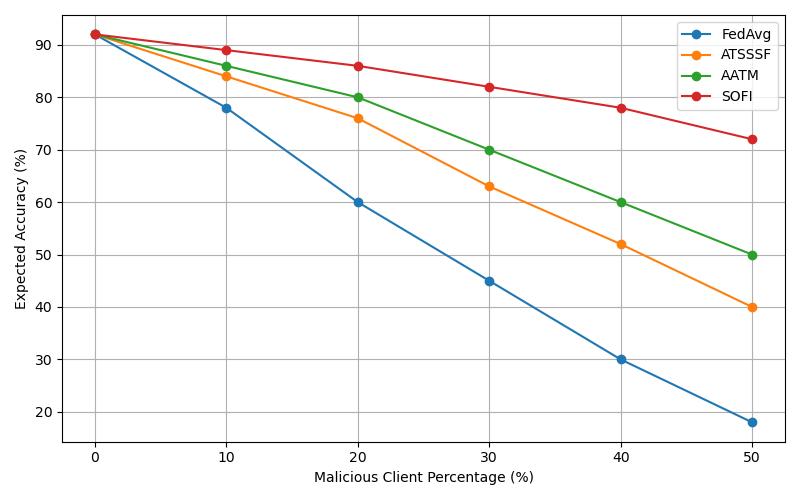}
    \caption{Expected Global Model Performance Under Increasing Attack Intensity}
    \label{fig:placeholder}
\end{figure}
Through its observation analysis decision action cycle, ATCL is expected to detect emerging instability earlier than threshold based mechanisms and apply targeted corrections such as threshold steering or smoothing adjustments before model degradation becomes significant. Because these actions are driven by contextual reasoning rather than static statistical rules, omission decisions should become more accurate, more situationally aware, and less sensitive to fixed hyperparameters.
\begin{figure}
    \centering
    \includegraphics[width=1\linewidth]{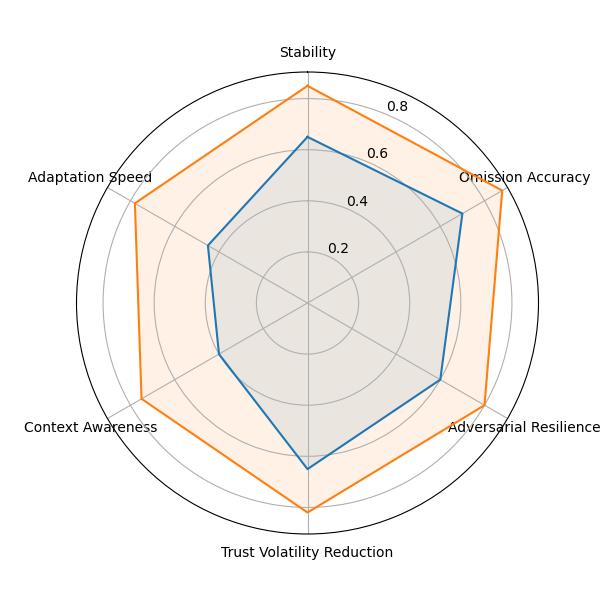}
    \caption{Expected improvement of ATCL over adaptive ATSSSF in terms of trust stability and omission accuracy}
    \label{fig:placeholder}
\end{figure}
The agentic nature of ATCL also supports more stable trust score evolution. Instead of responding mechanically to each deviation, the agent can interpret whether changes represent noise, transient effects, or sustained behavioural drift, enabling more deliberate and selective interventions. This behaviour is expected to reduce volatility in trust trajectories and mitigate oscillatory exclusion cycles often observed in static filters. Furthermore, the agent’s ability to integrate multiple behavioural cues update similarity, participation consistency, volatility indicators, and model level trends should improve detection of inconsistent or adversarial clients beyond what purely statistical rules can achieve.

Despite these enhancements, computational overhead remains low because the agent operates exclusively at the server using aggregated signals, without modifying client side execution. Taken together, the projected outcomes suggest that an agent driven trust mechanism such as ATCL can deliver greater robustness, stability, and interpretability in FL systems operating under heterogeneous and dynamic conditions.
\section{Conclusion and Future Developments}
This work introduces an agentic trust control loop (ATCL) for FL, extending earlier adaptive ATSSSF mechanisms into a general observation analysis decision action framework. By modelling trust relevant behavioural signals, applying lightweight reasoning, and issuing targeted interventions, ATCL provides a feasible and scalable path toward agentic trust coordination suited to large scale and multi domain FL deployments. Although still in an early stage of development, its projected benefits are grounded in prior adaptive ATSSSF results, which demonstrated improvements in model stability, omission accuracy, and resilience to unreliable or inconsistent clients. ATCL is expected to enhance these outcomes by enabling more context aware intervention and by preventing volatility caused by short term noise or adversarial perturbations.

The proposed trajectory of innovation follows a staged progression that begins with the practical implementation of ATCL. This first stage replaces fixed statistical thresholds with an agent driven trust controller capable of recognising instability, interpreting training state signals, and performing targeted trust adjustments entirely at the server side. Beyond this immediate contribution, the work outlines a longer term research direction toward Self Organising Federated Intelligence (SOFI): a broader, 6G aligned vision in which multiple coordinated agentic modules autonomously manage trust dynamics, model reliability, and system wide adaptation. While SOFI represents the architectural endpoint, ATCL provides the implementable foundation that enables such evolution.

Future work will focus on deploying and evaluating ATCL in realistic federated environments, assessing its operational overhead, its interaction with diverse models and client behaviours, and its integration requirements within existing FL platforms. This will establish the empirical validity of the proposed framework while guiding the gradual transition from agentic trust control toward fully self organising federated intelligence systems.
\section{Acknowledgments}
This work was supported by the HORIZON EUROPE Marie Sklodowska Curie Actions EXARCH project, under
Grant n. 101236244

\end{document}